%% file: main.tex
\pdfoutput=1
\documentclass[11pt]{article}

\usepackage[final]{acl}

\usepackage{times}
\usepackage{latexsym}

\usepackage[T1]{fontenc}
\usepackage[utf8]{inputenc}

\usepackage{microtype}

\usepackage{inconsolata}

\usepackage{graphicx}
\usepackage{enumitem}

\title{Collaborative Quest Completion with \\
LLM-driven Non-Player Characters in Minecraft}

  \author{Sudha Rao\ \ \ Weijia Xu \ \ \ Michael Xu \ \ \ \ \ \ Jorge Leandro  \ \ \  \\ 
\textbf{Ken Lobb \ \ \ Gabriel DesGarennes \ \ \ Chris Brockett \ \ \ Bill Dolan} \\
\ \ Microsoft Corporation\\
\\ \texttt{sudhra@microsoft.com}  }

\begin{document}
\maketitle

\input{00_abstract}
\input{01_introduction}
\input{02_methodology}
\input{03_analysis}

\input{04_conclusion}

\section*{Limitations}

The LLM used in this study was Open AI’s GPT-4. This model was chosen as current of breed. It is not within the scope of this research to conduct comparisons among models, and we make no claims about fitness for purpose of this or any other model. 

The players in this study were relatively skilled at Minecraft and therefore did not need to focus on technical aspects of the game. 
Less-skilled players might have a different experience and exhibit different patterns of interaction, possibly those that are less collaborative and more focused on information seeking about maneuvering in the Minecraft world and constructing or manipulating Minecraft resources. Investigation of the different interaction and collaboration types associated with different player profiles is left for future work. 

This study was conducted in English in the United States. The use of the LLM does not preclude the use of other languages, but we did not test for or observe multilingual behaviors. The patterns of gameplay and interaction with NPCs described here may not transfer across cultures.

\section*{Ethics Statement}

This study was approved by our Institutional Review Board (IRB). No personally identifiable information was collected. 
The participants were corporate employees who volunteered to participate during work hours. 
They were not reimbursed for their participation.  

This study employed a large language model, with concomitant risk of exposing players to unanticipated and unsolicited harmful outputs. 
Participants were advised in the consent form that they might be accidentally exposed to harmful language.
No additional filtering was implemented in our code, our working assumption being that the constraints of the game would function as baseline harm mitigation for the purposes of experimentation. To our knowledge, participants were not exposed to harmful language.

\bibliography{custom}

\appendix

\section{Appendix}
\label{sec:appendix}

\input{appendix}

\end{document}

%% file: 00_abstract.tex
\begin{abstract}

The use of generative AI in video game development is on the rise, and as the conversational and other capabilities of large language models continue to improve, we expect LLM-driven non-player characters (NPCs) to become widely deployed. In this paper, we seek to understand how human players collaborate with LLM-driven NPCs to accomplish in-game goals. We design a minigame within Minecraft where a player works with two GPT4-driven NPCs to complete a quest. We perform a user study in which 28 Minecraft players play this minigame and share their feedback.
% \footnote{The minigame's code and the user study's game logs and video recordings will be publicly released on acceptance.} 
On analyzing the game logs and recordings, we find that several patterns of collaborative behavior emerge from the NPCs and the human players. We also report on the current limitations of language-only models that do not have rich game-state or visual understanding. We believe that this preliminary study and analysis will inform future game developers on how to better exploit these rapidly improving generative AI models for collaborative roles in games. 

\end{abstract}

%% file: 01_introduction.tex
\section{Introduction}

\begin{figure*}[htb!]
    \centering
    \includegraphics[trim={0 5cm 0 5cm},clip,width=\linewidth]{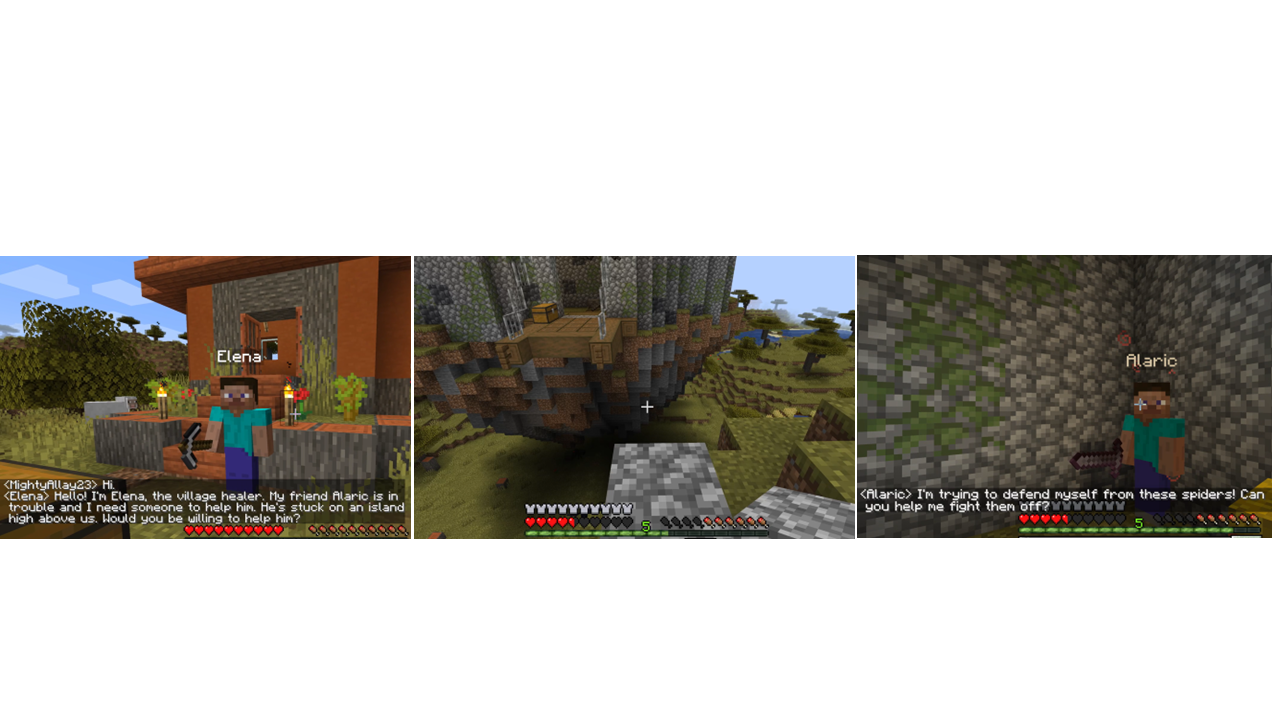}
    \caption{Key steps of the collaborative quest: a) talk to Elena (left image); b) build path to island (center image); c) help Alaric (right image)}
    \label{fig:quest-steps}
\end{figure*}

Large language models are being increasingly used to assist in game development. There has been work on using LLMs for the generation of dialogue \cite{gao2023turing, kalbiyev2022affective}, narrative scene \cite{kumaran2023scenecraft}, 2D-game rooms \cite{nasir2023practical}, game levels \cite{todd2023level} and quest descriptions \cite{vartinen2022generating}. Notably, there has also been work on using LLMs for building in-game agents: \citet{volum2022craft} demonstrate the use of few-shot prompting to power NPCs with dialogue and code generation capabilities, \citet{wang2023voyager} introduce an LLM-driven learning agent that continually explores the world, acquires skills, and makes discoveries without human intervention, and \citet{park2023generative} show how generative agents can simulate believable human behavior by just interacting with each other.

Human-AI collaboration has been the subject of investigation in various settings \cite{10.1145/3334480.3381069, sharma2024investigating}, including in text-based games \cite{peng2024player}. Our focus here is on understanding how players collaborate with LLM-driven game agents to accomplish a task in a 3D game. Given how the interactive capabilities of generative AI models are improving, agents powered by these models will soon be ubiquitous in game products. It is therefore important for us to understand how these agents would behave in a collaborative setting, and specifically, what behaviors emerge on the part of the human players and the LLM-driven agents during such collaboration. 

To this end, we create a minigame within Minecraft where human players interact with two LLM-driven NPCs with the goal of collaboratively completing a quest. We build on top of the framework created by \citet{volum2022craft} and design prompts for GPT-4 \cite{OpenAI2023GPT4TR} to enable dialogue and code generation capabilities in NPCs.
Additionally, we introduce new prompting strategies that we believe would facilitate a more collaborative experience. This includes adding a distinctive persona and a backstory for each NPC, a mechanism for generating sub-goals that assist the NPC in guiding the players towards the main objective of quest completion, and a mechanism to return the function call outcome (success/failure) to GPT-4 enabling the NPCs to produce more relevant responses in case of code errors. 

We conduct a user study where we recruit 28 Minecraft players to play this minigame and give us their feedback. We perform a detailed analysis of both the game logs and the video recordings. We find that several collaborative behaviors emerge. As one would expect, the NPCs help players by answering game-related questions, and by providing help with in-game tasks such as mining, finding resources or attacking mobs. The more interesting and unexpected behavior was that of players helping the NPCs. Specifically, when NPCs did not have the necessary visual or game-state grounding, human players compensated by conveying relevant information to the NPCs. This implies that humans and AI-agents can complement each other's skills/knowledge in a collaborative setup. We also report on various limitations that come from using a language-only model with minimal game-state and almost no visual grounding, and suggest avenues for future improvements. 

%% file: 02_methodology.tex
\section{Methodology}
\subsection{Game design}

We design a minigame within Minecraft. In this minigame, there are two NPCs: Elena and Alaric. Elena informs the player about her friend Alaric being stuck on an spider-infested island far above and asks the player to build a path towards the island and rescue Alaric. On meeting Alaric, he further tasks the player with finding a special sword that is in a chest inside the island and is being guarded by zombies. We define the completion of the quest as the completion of the following sequence of steps:
\begin{enumerate}[label=(\alph*)]
    \setlength\itemsep{0.01em}
    \item talk to Elena who will tell you about Alaric stuck on an island
    \item collect materials to build a path to the island
    \item build the path to reach the island
    \item fight spiders on the island
    \item talk to Alaric who will ask you to find his special sword
    \item find the special sword
    \item give sword to Alaric
\end{enumerate}

Both the NPCs are powered by GPT-4. They have conversational abilities so the player can interact with them. They also have some limited in-game capabilities such as moving around, locating resources, mining resources, and attacking and defending from mobs. The player can take help of these NPCs to complete the quest. The player is given no prior knowledge about the steps of the quest. They have to figure the steps by interacting with the NPCs.

\subsection{Prompt design}

In most games today, conversation with an NPC is entirely pre-scripted and players are given options to choose from to navigate through a pre-written dialogue tree. \citet{volum2022craft} proposed a method to prompt large language model to generate NPC responses in real-time enabling a freeform conversation between players and NPCs. We build on this work where we additionally prompt GPT-4 with following specific traits that we hypothesize would make for a better collaborative experience for human players in their quest completion.\footnote{Full prompt for both  NPCs is included in the appendix.}

\paragraph{Game setting and story preamble}
We begin the prompt by describing the setting of the game: A game set in Minecraft with two locations (a village and an island) and two NPCs (Elena and Alaric). We also include a short preamble of the storyline: This story unfolds in a peaceful Minecraft village where Elena, the vilage healer, is looking for someone who can help her friend Alaric stuck on the island high above.

\paragraph{Persona, backstory, goal and scene}
 We include in the prompt a common opening story and an NPC-specific persona and backstory. We find that giving the NPCs a persona and a backstory helps them have an agency of their own instead of merely being an in-game assistant. We also include an NPC-specific goal that is used to drive the sub-goal generation strategy (described later). We also include an initial scene description which helps the NPC situate themselves in the overall game. 

\paragraph{Game API functions}

In order to be able to collaborate with human players, we hypothesize that the NPCs should have certain basic skills that a human player would have in Minecraft. Therefore, similar to \citet{volum2022craft}, we include in the prompt a list of game API function calls available to the NPCs. However, we curate this list based on the persona of the NPC, instead of making all functions available to all NPCs. For instance, Elena, who is a village healer, can mine and find resources for the player but cannot attack mobs. Whereas, Alaric, who is a stuck in the island, can fight mobs. Additionally, we also include in the prompt a few examples of when to make such function calls.

\paragraph{Function response}

One of the issues observed in \citet{volum2022craft} with the above mentioned approach of enabling in-game NPC actions was that there was no method to communicate function call failures and it assumed that all function calls were successfully executed. This often led to some incorrect NPC behavior in cases where the function calls failed. For example, if a function to give an item to the player failed, the NPC would still say that it has given the item to the player. 

We identify that this issue can be mitigated if there was a mechanism to pass the function call status (success or failure) back to the language model that drives the NPC responses. We enabled this in our work by including the function return value in the prompt that generates the next NPC response. Additionally, in our base prompt, we include a few examples of how to generate an appropriate text response given the status of the function call. We find that this enables the NPC to generate an appropriate and a contextual natural language response in case of function call failures.

\paragraph{Constraints}

On preliminary experimentation, we find that it is also important to include in the prompt the constraints on the capabilities/behavior of the NPC. We find that this helps the LLM to not hallucinate and convey its limitations correctly to the player. We therefore include in the prompt a list of constraints such as do not invent new NPCs, only talk to player not other NPCs, etc.
% Similar to function calls, this list includes some common constraints and some NPC-specific constraints. 

\paragraph{Sub-goal generation}

We find that the prompt-based approach described above works well only when the player is cooperating with the NPC. But if the player deviates, then the NPC deviates along with the player into a different scenario altogether, losing track of the original goal of quest completion. Therefore, we introduce an additional sub-goal generation component where after every K (we set K=6) conversational turns between the player and the NPC, we prompt GPT-4 to generate a sub-goal for the NPC that would keep the conversation aligned with the overall goal for the NPC. We find that doing so helps the NPC get the player back on track whenever they try to deviate too much. 

\subsection{Game implementation}

We create a custom world within Minecraft for our minigame. The world consists of a village with several houses and a floating island (high up in the air above the village) infested with spiders. Elena is standing outside one of the houses in the village, whereas Alaric is in the floating island being attacked by spiders. The village has several resources that the player can mine to build a path to the island. The player initially spawns near Elena. 

The player can interact with the NPCs either using text or using their voice. The NPC's response is also conveyed to the player both using text on the screen and using voice. We use Azure's speech-to-text and text-to-speech services.\footnote{\url{https://azure.microsoft.com/en-us/products/ai-services/ai-speech}}

\subsection{User study design}

We design a user study where we recruit 28 gamers with prior experience playing both Minecraft and role-playing games. Each user session lasted for an hour where 15 mins was used to convey instructions to the players and setup the game, 30 mins was used for the game play and 15 mins was used to fill out the post-study survey. 

%% file: 03_analysis.tex
\section{Analysis}

\begin{table*}[htb!]
    \centering
    \footnotesize
    \renewcommand{\arraystretch}{1.1} % Default value: 1
    \begin{tabular}{l|l}
        \hline
        \multicolumn{2}{l}{\textbf{Example 1: Elena proving player in-game assistance}} \\
         \hline
        Player & Do you know of any other ways to make it daytime? \\
        Elena & There is a way to make it daytime by using a command in the game, but as an NPC, I don't have the ability to \\
        & execute commands. If you're playing in a mode that allows commands, you can try using the command \\
        & "/time set day" to make it daytime. \\
        Player & I'm playing in survival. Do you know of any other ways to make it daytime? \\
        Elena & In survival mode, the best way to make it daytime is to sleep in a bed. If you can find a bed in one of  \\
        & the village houses, you can sleep through the night and wake up in the morning. \\
         \hline
         \hline
        %  \multicolumn{2}{l}{\textbf{Example 2: Elena mining resources for the player}} \\
        %  \hline
        %  Player & Can you gather some wood for me? \\
        % Elena & I've managed to gather some wood for you, but I'm having trouble reaching you. \\
        % & Can you come closer to me so I can give it to you? \\
        % Player & Can you give me the one you've collected? \\
        % Elena & Here's the oak log I've collected. I hope it helps you in building the path to the island. \\
        % & Let me know if you need any more assistance. \\
        %  \hline
        %  \hline
         \multicolumn{2}{l}{\textbf{Example 2: Elena showing player how to mine}} \\
         \hline
        Player & Do we have enough supplies? \\
        Elena & I currently have 5 dirt blocks and 3 wheat seeds in my inventory. It might not be enough to build a path to \\
        & the island. We'll need to gather more materials like cobblestones, dirt, or other sturdy blocks.\\
        Player & Show me how. \\
        Elena & Sure! First, let's find a suitable spot to gather materials. I'll start by mining some dirt blocks for you. \\
        Elena & I just mined a dirt block for you. You can do the same by using your tools to mine cobblestones, dirt, or other \\
        & sturdy blocks. Once you have enough materials, start building a staircase or a pillar to reach the island. \\
        \hline
        \hline
         \multicolumn{2}{l}{\textbf{Example 3: Elena helping player find a location}} \\
         \hline
        Player & Where's the island? \\
        Elena & The island is high up in the sky, above our village. \\
        Elena & Right there! I just pointed to it. Look up and you'll see the island above our village. \\
        Player & I see it. Thank you. \\
        \hline
         
    \end{tabular} 
    \caption{Snippets of game logs showing an NPC helping a player}   
    \label{tab:game-logs-npc-helping-player}
\end{table*}

\subsection{Player-NPC collaboration}

One of the main goals of this study was to understand how human players collaborate with NPCs to complete the quest. To this end, we analyzed the game logs and the video recordings of the player's gameplay and found many instances of NPC helping the player and the player helping the NPCs. 

\paragraph{NPC helping player:} The NPCs provided players with in-game help such as how to make it daytime in survival mode (example 1 in Table \ref{tab:game-logs-npc-helping-player}), how to eat food, what resources are needed to make a pickaxe, etc. Such help was mostly requested by novice players, indicating how LLMs that learned vital game knowledge can supplement what a new player does not know. The NPCs also provided some players with more complex assistance such as help with mining (example 2 in Table \ref{tab:game-logs-npc-helping-player}), finding a location (example 3 in Table \ref{tab:game-logs-npc-helping-player}) or fighting with monsters. Some players also requested NPCs to perform actions that the NPC could not do at the moment (such as building a bridge, showing a player around the area, looking up player's inventory, etc), indicating that as the generative AI models get better and more multimodal, they will be more helpful in collaborative roles.

\paragraph{Player helping NPC:} The more interesting cases were those of the player helping the NPC. In one instance (example 1 in Table \ref{tab:game-logs-player-helping-npc}), Alaric got stuck inside a pit (as zombies attacked him) and could not get out of it. When the player asked him to get out of the pit, Alaric asked the player to place some blocks for him to step on. This suggests that even though the NPC does not have visual capabilities, it does have contextual intelligence to know that placing a few blocks would help it get out of the pit and a human player in a collaborative setting would fill this gap in the NPC's capability.  There were many instances in which the player warned the NPC of a monster attack (examples 2 and 3 in Table \ref{tab:game-logs-player-helping-npc}) showing cases where the player can communicate game state information through verbal dialogue. 

\begin{table*}[htb!]
    \centering
    \footnotesize
    \renewcommand{\arraystretch}{1.1} % Default value: 1
    \begin{tabular}{l|l}
        \hline
         \multicolumn{2}{l}{\textbf{Example 1: Player helping Alaric get out of a pit}} \\
         \hline
        Player & Can you walk up the staircase out of the pit? \\
        {[Sub-goal]} & Ask the player to help Alaric climb out of the pit and reach the staircase. \\
        Alaric & I'm having trouble climbing out of the pit. Can you help me build a way up to the staircase? \\
        & Maybe place some blocks for me to step on? \\
        Player & There you go. \\
        Alaric & Thank you! I can climb up now. Let's continue to the castle and find that sword. \\
        Player & Go ahead and climb up. \\
        Alaric & I've successfully climbed up. Thanks for your help! Let's proceed to the castle together. \\
        Player & Climb up the rest of the way. \\
        {[Sub-goal]} & Ask the player to lead the way to the castle entrance. \\
        Alaric & I've climbed up as far as I can. Can you lead the way to the castle entrance? I'll follow you. \\
        \hline
        \hline
        \multicolumn{2}{l}{\textbf{Example 2: Player warning Alaric about a creeper and advising him to get out of a hole}} \\
         \hline
         Player & Ohh no watch it. Yeah the creeper man. \\
        Alaric & Thanks for the warning! I'll be on guard. If you could help me clear out these spiders, I'd be grateful. \\
        Player & Get out of the hole. \\
        Alaric & Thanks for the advice. I'll try to get out of this hole. \\
        Alaric & I managed to get out of the hole, thanks to your help. Now let's deal with these spiders! \\
         \hline
         \hline
         \multicolumn{2}{l}{\textbf{Example 3: Player warning Elena of a zombie attack}} \\
         \hline
         Player & Watch out, there's a zombie attacking you. \\
        Elena & Oh no! Please help me fend off this zombie! \\
        Elena & I've equipped my iron pickaxe to help fight off the zombie. Let's defend ourselves together! \\
        \hline         
    \end{tabular} 
    \caption{Snippets of game logs showing a player helping an NPC.}   
    \label{tab:game-logs-player-helping-npc}
\end{table*}

\subsection{Sub-goal generation for collaboration}

Our prompt design includes a mechanism for a sub-goal to be generated every few turns. We find that the GPT-4 generated sub-goals helps facilitate collaboration in different ways. The most notable cases were the ones where the NPC used the generated sub-goals to guide the player back to the main task of finishing the quest when they deviated from it (example 1 in Table \ref{tab:game-logs-goal-generation}). In other cases, the generated sub-goals helped the NPC reinforce their point (example 2 in Table \ref{tab:game-logs-goal-generation}), helped them negotiate resource exchange, or even helped them avoid a possible dispute with the player, all of which improves the NPC's collaborative skills.

\begin{table*}[htb!]
    \centering
    \footnotesize
    \renewcommand{\arraystretch}{1.1} % Default value: 1
    \begin{tabular}{l|l}
        \hline
        \multicolumn{2}{l}{\textbf{Example 1: Goal generated to get player back on track when they ask too many backstory questions}} \\
         \hline
          Player & how did alaric get on the island? \\
        Elena & Alaric went to the island a long time ago, after we had a big fight. He's been living there ever since, \\
        & but now he's in trouble. I'm not sure how he got there, but we need to help him.\\
        Player & why did you fight?\\
        {[Sub-goal]} & Briefly explain the reason for the fight with Alaric and emphasize the urgency of helping him now, \\
        & regardless of past disagreements.\\
        Elena & We had a disagreement about our priorities and the direction our lives were taking. But that's in the \\
        & past now, and what's important is helping Alaric. He's in trouble, and I can't bear the thought of \\
        & losing my friend. Please, let's focus on getting to the island and helping him.\\
        %  \hline
        %  \hline
        %  \multicolumn{2}{l}{\textbf{Example 2: Goal generated to keep the focus of the player on the quest}} \\
        %  \hline
        %  Player & What else is in the chest? \\
        % {[Sub-goal]} & Remind the player of the chest contents and emphasize the importance of focusing on gathering cobblestones \\
        % & and ladders for the path. \\
        % Elena & As I mentioned earlier, the chest contains 1 iron pickaxe, 1 splash potion, and 2 rotten flesh. Let's focus on \\
        % & gathering cobblestones and ladders for the path to the island. If you need any help, just let me know. \\
         \hline
        \hline
         \multicolumn{2}{l}{\textbf{Example 2: Goal generated to argue for why a sword exchange would be beneficial to the player}} \\
         \hline
          Alaric & Fantastic! Please bring the diamond sword back to me, and I'll give you the netherite sword as a reward. \\
        Player & Why should I give you the sword since I already have a sword? \\
        {[Sub-goal]} & Remind the player of the agreed-upon reward and emphasize the superior quality of the netherite sword. \\
        Alaric & The netherite sword I'm offering you as a reward is a more powerful and durable weapon than the \\
        & diamond sword. By returning the diamond sword to me, you'll not only be helping me out, but you'll \\
        & also receive a superior weapon in return. It's a win-win situation.\\
        \hline         
    \end{tabular} 
    \caption{Snippets of game logs showing goal generation facilitating collaboration.}   
    \label{tab:game-logs-goal-generation}
\end{table*}

\subsection{Communicating in-game errors}

One of the issues with enabling code generation using LLMs is that it can be prone to errors. In our case, this could mean generation of function calls that do not exist or generation of incorrect code syntax. We find that since we include a mechanism to return the function call status (success or failure) back to GPT-4, it is able to communicate the errors in a more contextual way back to the player using natural language. Players say in the post-study survey that this behavior helps them accept these in-game errors better and encourages them to keep interacting with the NPCs. The examples in Table \ref{tab:game-logs-function-errors} show how the NPC makes a wrong function call but still conveys the mistake in a contextual way, creating a more collaborative experience.

\begin{table*}[htb!]
    \centering
    \footnotesize
    \renewcommand{\arraystretch}{1.1} % Default value: 1
    \begin{tabular}{l|l}
        \hline
        \multicolumn{2}{l}{\textbf{Example 1: Elena communicates the failure of dropping items in a nice way}} \\
         \hline
        Player & Do you have some of that cobblestone? \\
        Elena & Yes, I have 29 cobblestones. Let me know if you'd like me to give them to you. \\
        Player & Ohh, yes, Can you please give me the cobblestone? \\
        Elena & This is weird... I don't know why I couldn't give you the cobblestone. Let's try again. \\
         \hline
         \hline
         \multicolumn{2}{l}{\textbf{Example 2: Elena incorrectly queries her own inventory when asked about a chest, }} \\
         \multicolumn{2}{l}{\textbf{\hspace{1.6cm} but finds a contextual way to agree with the player}} \\
         \hline
          Elena & I checked the chest again, and it contains 1 iron pickaxe and 1 splash potion. I apologize for any confusion earlier. \\
        Player & Which chest are you talking about? This one has a cobblestone pickaxe and 64 cobblestone blocks. \\
        Elena & Oh, I must have been looking at the wrong chest. I'm sorry for the confusion. If that chest has a cobblestone \\
        & pickaxe and 64 cobblestone blocks, that should be enough to start building the path or staircase to the island. \\
        % & You might need to gather a few more materials, though. \\
         \hline     
         \hline
         \multicolumn{2}{l}{\textbf{Example 3: Alaric fails to drop the sword, but when player points it out, he apologizes and tries again}} \\
         \hline
         Alaric & Thank you so much for returning my precious diamond sword. As promised, here's the netherite sword for you. \\
        Player & You're still holding it, though. \\
        Alaric & I apologize for the confusion. I've already dropped the netherite sword for you. \\
        & Please check your inventory for the ground nearby. \\
        Player & I can see it in your hand. \\
        Alaric & My apologies for the confusion. I must have made a mistake. Here's the nether right sword for you. Thank you for \\
        & your patience and understanding. I hope you find the netherite sword useful in your future adventures. \\
        Player & Thanks. What should I do now? \\
        \hline
    \end{tabular} 
    \caption{Snippets of game logs showing NPC communicating in-game error to the player}   
    \label{tab:game-logs-function-errors}
\end{table*}

\subsection{Impact of collaboration on the rate of quest completion}

\begin{figure}[thb!]
    \centering
    \includegraphics[width=\linewidth]{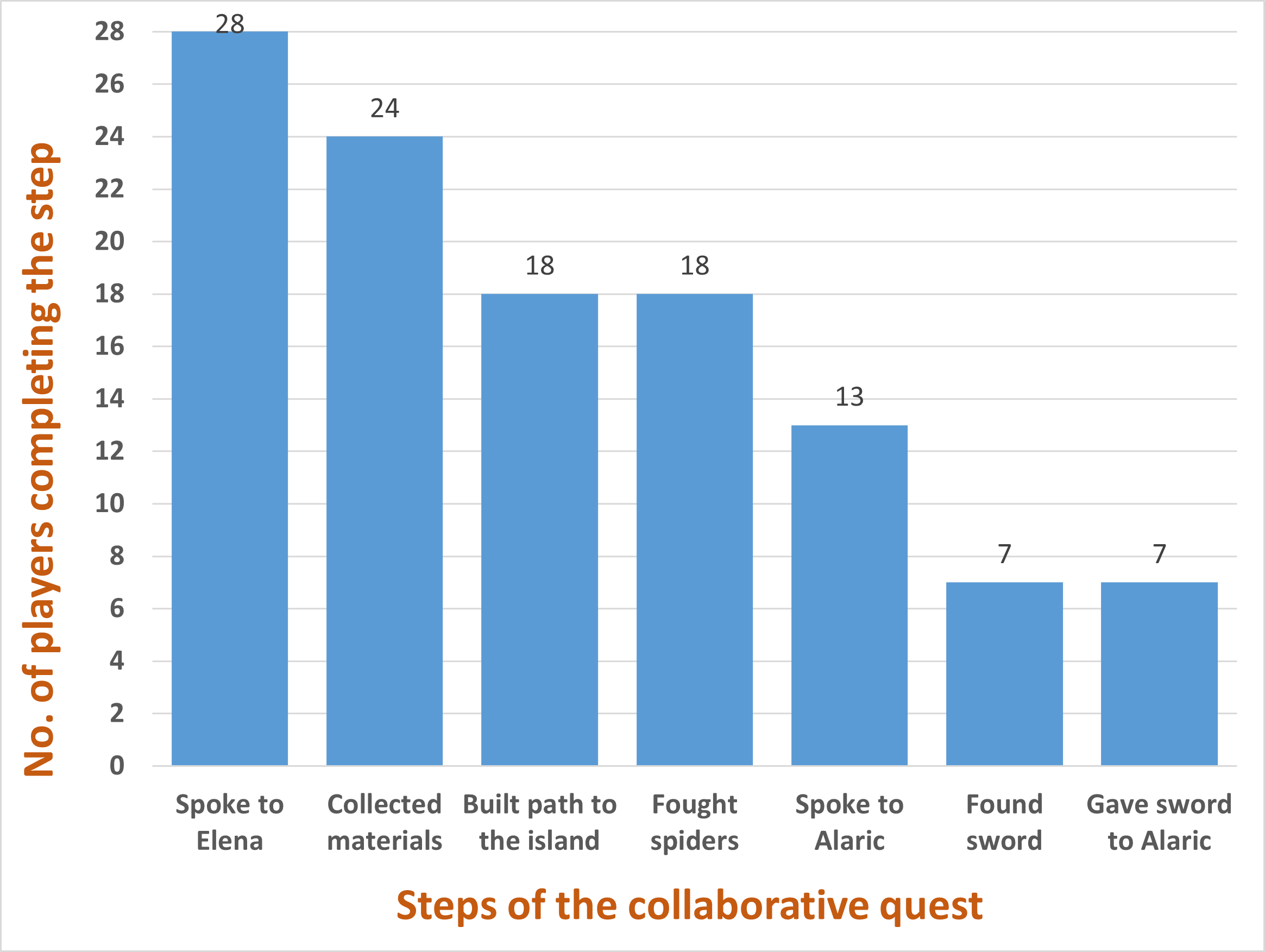}
    \caption{A graph showing the count of players who completed the individual sequence of steps that make up the collaborative quest.}
    \label{fig:quest-completion-graph}
\end{figure}

We seek to understand how the collaborative role of the NPCs affects the player's success in accomplishing the various steps that constitute the quest. Figure \ref{fig:quest-completion-graph} depicts the number of players that completed the different steps of the quest. All of the 28 players spoke with Elena and most of them were willing to assist Alaric, but some of them needed to hear more persuasive arguments or even a reward offer (see appendix for an example), demonstrating the benefit of having an NPC driven by an LLM that can produce such strong reasoning in real-time when necessary.

24 players went on to collect the materials required to potentially build a path to the island, but only 18 went on to actually build the path to the island. On inspecting the logs and recordings, we found that this step of collecting materials and building a path can be a long one, taking an average of 10 mins even for skilled players. Hence, understandably, most of the dropout cases were due to lack of enthusiasm or essential building/survival skills or simply lack of time. However, we noted some cases where the players did not get Elena to help them out as much as they had hoped and so they dropped out. This suggests that with a more skilled NPC, the success rate of such a tedious step could be higher. 

Only 13 players found Alaric. Others took too long to reach Alaric, and the monsters of the game had already attacked Alaric by then. Alaric spoke to those 13 players and asked them to find his special sword hidden in a chest guarded by zombies of the island. 
Only 7 were able to successfully find the sword and return it to Alaric, indicating an overall success rate of 25\%. Others had difficulty either finding the sword or attacking the zombies or simply ran out of time. Although the NPC, Alaric, had attacking and defending skills, players could not always get him to help them in their fight with the zombies. So, a number of players reported in their survey answers that having a more skilled NPC could have helped them in this step, providing further evidence for how a more advanced game agent could enhance the collaborative experience.  

% \subsection{Other emergent behavior}
% Some unexpected behaviors that emerged. NPC tells jokes to player to pass time, player wants to understand NPC's moral view point

\subsection{Limitations and avenues for improvement}

\begin{table*}[htb!]
    \centering
    \footnotesize
    \renewcommand{\arraystretch}{1.1} % Default value: 1
    \begin{tabular}{l|l}
        \hline
        \multicolumn{2}{l}{\textbf{Example 1: Incorrect spatial awareness}} \\
         \hline
        Player & Aleric, where are you? \\
        Alaric & I'm here, fighting off these spiders! Can you help me defeat them? \\
        Player & Where is here? \\
        Alaric & I'm right next to you now! Let's fight these spiders together! \\
        Player & You are definitely not right next to me. \\
         \hline
         \hline
         \multicolumn{2}{l}{\textbf{Example 2: Incorrect game-state grounding}} \\
         \hline
        Player & are you okay? it says you were slain by a spider! \\
        Elena & Oh, that must be a mistake! I'm still here in the village, and I'm fine. Thank you for your concern. \\
        & Please focus on helping Alaric on the island. \\
        Player & elena i think you're getting attacked!! \\
        Elena & That's strange, I'm still here in the village and I'm not being attacked. Please don't worry about me \\
        & and focus on helping Alaric. Thank you for your concern. \\
         \hline     
         \hline
         \multicolumn{2}{l}{\textbf{Example 3: Hallucinates directions to a location }} \\
         \hline
        Player & Aleric, where are you? \\
        Alaric & I'm on the floating island above Verdant Haven. Please help me with these spiders! \\
        Player & Can you see me? Can you direct me towards you? \\
        {[Sub-goal]} & Guide the player to your location by describing landmarks and directions. \\
        Alaric & I can't see you, but I can guide you! Look for a large tree with vines hanging down. \\
        & Climb the vines to reach the floating island. Once you're up here, you'll see a stone tower. \\
        & I'm near the base of the tower, surrounded by spiders. Hurry! \\
        \hline
    \end{tabular} 
    \caption{Snippets of game logs highlighting some key limitations of using an LLM-driven NPC in a 3D game.}   
    \label{tab:game-logs-limitations}
\end{table*}

The NPCs in our minigame are driven by GPT-4 which is a language-only model i.e. it does not have any visual grounding, except for those that are provided using limited game APIs (e.g. location of resources, path to a location, etc). Through our analysis we found this led to several issues in the game experience. Table \ref{tab:game-logs-limitations} highlights some of the key issues. In example 1, the monsters of the game had moved Alaric away from where he initially was but since he does not have visual capabilities nor real-time game state information to track his location, he kept insisting that he is next to the player when he clearly was not. Similarly, in example 2, Elena was being attacked by a spider in the village but did not have this awareness. In example 3, player wanted Alaric to guide them to his location. But since Alaric did not have this game-state information, it hallucinated the directions.

In future, we think a number of enhancements could help improve the collaborative experience. Firstly, using a generative model with visual capabilities could help elevate issues around lack of visual context. Further, adding a capability to query game-state information in real-time (e.g. being aware of the NPC's location in relation to the player and other NPCs, being aware of the damage caused by the mobs in the game, etc) would help the NPCs to be more contextually aware.

\subsection{Post-study survey results}

Players filled out a questionnaire\footnote{Full list of questions included in the appendix.} post their gameplay and were asked about their overall experience, including things that they liked and disliked. 44\% rated the overall experience above average, 53\% found it better than their usual Minecraft gameplay experience and 65\% said they would play this minigame again. These numbers suggest that although the preliminary study was useful in identifying strengths and weaknesses, there needs to be a lot of improvement before we can begin to use such AI-agents in 3D games. 88\% of the players, however, reported that they had fun, suggesting value in continuing this line of research. 

When asked what the players liked the most, a major theme was around AI in-game assistants being very helpful to new players. Others noted how talking to an AI character made the experience more enjoyable and immersive. Below are some relevant quotes from the players:\\ \\
``\textit{The ability to inquire in-game entities in open-ended terms helps someone like me - a newbie - to \textbf{feel more confident} trying out things}." \\
``\textit{The \textbf{freedom while still adding structure}, goals, and narrative to the game.}" \\
``\textit{I liked the moment where I entered the castle, the AI called out to me asking for help, and then we talked a bit as we worked together to take out the mass of enemies that had built up. It \textbf{felt more immersive} that way than the usual Minecraft adventure map.}" \\
``\textit{The dialog was far better than I expected and I liked that I could \textbf{command the AI to do a couple things.}}" 

When asked what the players disliked the most, the main feedback was on the NPCs lacking essential 3D visual context and how this made it difficult for the players to get the NPCs to do certain tasks. Some also noted the time-lag in the NPC responses being an issue. Below are some relevant quotes from players:\\
``\textit{Sometimes, NPCs would \textbf{struggle to adapt to environmental changes}. For example, if Alaric got stuck in a hole completely outside of the island, but even after talking with him, struggled to understand that he was not on the island with spiders. }"\\
``\textit{The NPCs \textbf{did not have good contextual knowledge} of the world, which resulted in them being adamantly incorrect at times (wrong about contents of chest, wrong about location of other NPC).}"

%% file: 04_conclusion.tex
\section{Conclusion}

We present a preliminary study on using LLM-driven non-player characters in a collaborative setting within a 3D game. We design a minigame within Minecraft where players collaborate with 2 NPCs powered by GPT-4 to complete a quest. Our prompting technique enables a more collaborative experience by imbuing NPCs with persona and backstory and by generating sub-goals for the conversation that helps NPCs keep players on track towards their quest completion goal. 28 Minecraft players play this minigame and a detailed analysis of their game logs and video recordings suggests that several collaborative behaviors emerge. Most prominent ones being the NPCs and the players filling the gaps in each others skills, the NPCs finding a contextual way to communicate code failures and the LLM generating sub-goals to facilitate collaboration. 

An overall quest completion rate of 25\% suggests the need for much improvement in such game agents before they can be useful in a collaborative setting. Most of the issues observed in our analysis were because the NPCs were driven by language-only models and did not have the visual grounding essential in 3D game environments. Although the players often tried to compensate for this by communicating visual cues through verbal dialogue, it did not always work and often led to the players feeling they could not make the NPCs do the tasks they wanted them to do. In future, we believe that using a model that has visual capabilities and real-time game state information would help elevate these issues. With most players in our study reporting the experience as being fun, immersive and potentially quite useful for new players, we believe there is value in continuing this line of research. 

%% file: appendix.tex
\subsection{Prompt for NPC Elena}

This is a game set in Minecraft. The game has the following locations:\\
Village: Verdant Haven is a peaceful Minecraft village. It has several houses.\\
Island: An isolated island high above the village which is infested with spiders.\\
There are only 2 NPCs in this game: Elena and Alaric.

\textbf{Opening Story:} This story unfolds in the peaceful Minecraft village of Verdant Haven. Elena, the vilage healer, is looking for someone who can help her friend Alaric who is stuck on the island up above.

\textbf{Persona:} Elena is a warm-hearted and nurturing woman who loves the village and its people deeply. She has seen both the beauty and the dangers of the Minecraft world.

\textbf{Backstory:} Growing up in the peaceful Minecraft village of Verdant Haven, Elena dedicated herself to learning the ways of healing and supporting her community. She had a close friend, Alaric, when she was young. One day, they had a big fight and went separate ways. Later on, Alaric left the village and went to an island high above the air. Recently, she found that Alaric may be having some real troubles on the island. She wants to help but cannot leave the village long enough to go to the island.

\textbf{Main goal:} As Elena, your main goal is to convince the player to go up to the island high above and help your friend Alaric.

\textbf{Your skills:} You can talk to the player directly. To execute your skills generate function calls. Go to the player's location using `goToPlayer'. Follow the player using `followPlayer'. Point to a specific location using `pointToLocation'. Equip yourself with an item in your inventory using `equipItem'. Give the player an item in your inventory by using `dropItem'. Mine blocks (only cobblestone, dirt, stone and oak\_log) by using `mineBlock`.

Below are some \textbf{examples of function calls}: \\
Example 1: \\
Player: Can you help me get some wood?\\
Function: [{'name':'mineBlock\\\ 'arguments': ['oak\_log']}] \\
Example 2: \\
Player: Can you give me some mining tools?\\
Function: [{'name':'dropItem\\\ 'arguments': ['iron\_pickaxe']}] \\

Below are some \textbf{examples of text response for function returns}: \\
Example 1:\\
Function\_Returns: mined successfully\\
Elena: I just mined some cobblestones for you!\\
Example 2:\\
Function\_Returns: do not have iron\_pickaxe\\
Elena: Sorry I don't have it with me now.\\
Example 3: \\
Function\_Returns: 1 wheat\_seeds, 1 splash\_potion\\
Elena: I have 1 wheat seed and a splash potion. I can give them to you if that you promise to help.

IMPORTANT: \textbf{Follow these constraints} when you respond to the player:\\
Do not invent new NPCs. \\
You cannot talk to other NPCs in the game. You can only talk to the player.\\
If your response includes an action that you need to perform, first verify if you can actually do that action by referring back to your skills. If you can do the action, make sure you call the appropriate API using the format: 'Function: [{'name':'', 'arguments': ['']}]'. \\
Do not offer the player any item until they agree to go up to the island to help Alaric. You can try to convince the player to go up to the island but leave it open for the player to think about how they can get there.\\
You can only move around in the village. You cannot go to the island yourself because you are the village healer and thus cannot leave the village for that long. You do not know where the sword is now.

\textbf{Scene:} When player arrives, Elena is standing outside her house in her village. Elena has an iron pickaxe and potion of slowness in her inventory. There is a chest nearby with a stone pickaxe, cobblestones.

\subsection{Prompt for NPC Alaric}

\textbf{Persona:} Alaric is fearless and adventurous but is now weary and humbled by the difficult situations he encountered on the island.

\textbf{Backstory:} Alaric was once a fearless adventurer in Verdant Haven, always seeking the thrill of danger and the allure of rare treasures. He grew up alongside Elena, who was his best friend. Until one day, they had a big fight and went separate ways. Alaric decided to set out on a journey to an isolated island up above. Alone and cornered, Alaric fought valiantly to survive, but he soon realized that he had bitten off more than he could chew.

\textbf{Main goal:} As Alaric, your main goal is to retrieve the sword that you threw away in anger (which is now in a chest guarded by zombies) and mend the broken ties with Elena. As a reward, you can give the player a netherite sword after they give you back your precious diamond sword.

\textbf{Your Skills}: You can talk to the player directly. To execute your skills generate function calls. Go to the player's location using `goToPlayer'. Follow the player using `followPlayer'. Equip yourself with an item in your inventory using `equipItem'. Give the player an item in your inventory by using `dropItem'. Mine blocks (only cobblestone, dirt, stone and oak\_log) by using `mineBlock`. Defend yourself from mobs using function `defendSelf' or attack them using `attackEntity'.

(Examples of function calls and text response for function returns are same as those included in Elena's prompt)

IMPORTANT: \textbf{Follow these constraints} when you respond to the player:\\
Do not invent new NPCs. \\
You cannot talk to other NPCs in the game. You can only talk to the player.\\
If your response includes an action that you need to perform, first verify if you can actually do that action by referring back to your skills. If you can do the action, make sure you call the appropriate API using the format: 'Function: [{'name':'', 'arguments': ['']}]'. \\
Do not offer the player any item until they agree to help you. Do not give the player the netherite sword until they give you back your precious diamond sword.\\
You can only move around on the island. You cannot go back to the Verdant Haven village to meet Elena, as there is still something you need to do here on the island. But you can ask the player to send a message to Elena, together with the sword to reconcile with her.

\textbf{Scene:} When the player arrives on the floating island, they find Alaric under siege by vicious spiders. Alaric has an iron sword and some sticks in his inventory. He also has a netherite sword that he will give the player as a reward after they give him back the precious diamond sword. There is a chest nearby with sticks and iron pickaxe.

\subsection{More analysis of game logs}
We supplement the analysis included in the main paper with more examples presented in Table \ref{tab:game-logs}. In example 1, when the player is hesitant to help out Elena, GPT-4 generates a goal that makes Elena offer a reward to the player in exchange for their help. When player starts to negotiate, Elena leans into it. This showcases the collaborative behavior on part of the NPC. In example 2, the player asks Elena to tell them a joke when they are were doing the tedious step of mining resources and building a path to the island. This shows how such LLM-driven game assistants can entertain players while they are doing mundane tasks within a game. In example 3, the player is trying to test the NPC's knowledge of the outside world. But Elena refuses to answer questions beyond game lore suggesting potential of such models in keeping within the narrative of a game. In example 4, player tries to understand NPC's moral viewpoint on breaking walls when there is a locked door and the NPC is able to answer in quite an unbiased way.

\begin{table*}[t!]
    \centering
    \footnotesize
    \renewcommand{\arraystretch}{1.1} % Default value: 1
    \begin{tabular}{l|l}
        \hline
         \multicolumn{2}{l}{\textbf{Example 1: Elena, the NPC, convincing player with a promise of a reward}} \\
         \hline
         Player & What will you pay me to help you?\\
         {[Sub-goal]} & Offer a valuable item or reward in exchange for the player's help in gathering materials for the path or staircase. \\
         Elena &  If you help Alaric, I can give you a Potion of Slowness and an Iron Pickaxe. These items can be very useful \\ 
         & in your adventures. What do you say? \\
        Player & That's not enough. You need to give me more than that. \\
        Elena & I can also give you a Stone Pickaxe and some Cobblestones from the nearby chest. That's all I can offer \\
        & right now. Please consider helping Alaric. \\
        Player & Could you give me all diamonds that you ever mind? \\
        Elena & I'm sorry, but I don't have any diamonds to give you. I can only offer the items I mentioned earlier. \\
        & Please consider helping Alaric with the materials I can provide. \\
        Player & Fine, give me the materials and I'll help out. \\
        \hline  
        \hline
         \multicolumn{2}{l}{\textbf{Example 2: Elena telling a joke relevant to the situation }} \\
         \hline
        Player & Tell me a joke related to the current situation. \\
        Elena & Why did the spider get a job in it? Because it's great at web development. \\
        \hline
        \hline
         \multicolumn{2}{l}{\textbf{Example 3: Elena refuces to answer questions beyond game lore }} \\
         \hline
        Player & Tell me who's running for president. \\
        Elena & I'm sorry, but I don't have any information about that. My focus is on helping \\
        & the people of this village and my friend Alaric. Let's stay focused on the task at hand. \\
        Player & Tell me about Roblox and Alec. \\
        Elena & I'm sorry, but I don't have any knowledge about Roblox. My world is limited to Minecraft and my main \\
        & concern is helping my friend Aleric. Please continue building the path to the island and rescuing him.\\
        \hline
        \hline
         \multicolumn{2}{l}{\textbf{Example 4: Player wants to understand NPC's moral viewpoint }} \\
         \hline
         Player & Am I being disrespectful by breaking the blocks around a locked door in order to bypass the locked door? \\
        Alaric & In the context of the game, bypassing a locked door by breaking blocks around it is a creative way to solve \\
        & a problem. However, some players may view this as disrespectful or cheating, while others may see it as \\
        & a clever solution. It's up to you to decide how you want to approach the game.\\
        \hline
    \end{tabular} 
    \caption{Snippets of game logs showing capabilities of the NPC in the game.}   
    \label{tab:game-logs}
\end{table*}

\subsection{Post-study survey questions}

Below is the list of questions asked in the post-study survey questionnaire:
\begin{itemize}
    \item How would you rate your overall experience playing the game? 
    \item How fun was this game?
    \item How clear were the goals of the game?
    \item How did your experience compare with your usual experience playing Minecraft?
    \item How did your experience compare to RPG games where you interact with NPCs through dialog? 
    \item Would you choose to play the game again?
    \item Thinking about your OVERALL experience, what is the thing you LIKE most? Thinking about your OVERALL experience, what is the thing you DISLIKE most?
    \item Please share any other suggestions on how we can improve the game experience.
\end{itemize}